\documentclass[a4paper]{article}

\usepackage[pdftex]{graphicx}
\usepackage{url}
\usepackage{float}
\usepackage{INTERSPEECH2018}
\DeclareMathOperator{\EX}{\mathbb{E}}% expected value

\title{Expressive Speech Synthesis via Modeling Expressions \\ with Variational Autoencoder}
\name{Kei Akuzawa$^1$, Yusuke Iwasawa$^1$, Yutaka Matsuo$^1$}
\address{
  $^1$Graduate School of Engineering, The University of Tokyo, Japan}
\email{\{akuzawa-kei,iwasawa,matsuo\}@weblab.t.u-tokyo.ac.jp}

\begin{document}

\maketitle
\begin{abstract}
Recent advances in neural autoregressive models have improve the performance of speech synthesis (SS).
However, as they lack the ability to model global characteristics of speech (such as speaker individualities or speaking styles), particularly when these characteristics have not been labeled, making neural autoregressive SS systems more expressive is still an open issue.
In this paper, we propose to combine VoiceLoop, an autoregressive SS model, with Variational Autoencoder (VAE).
This approach, unlike traditional autoregressive SS systems, uses VAE to model the global characteristics explicitly, enabling the expressiveness of the synthesized speech to be controlled in an unsupervised manner.
Experiments using the VCTK and Blizzard2012 datasets show the VAE helps VoiceLoop to generate higher quality speech and to control the expressions in its synthesized speech by incorporating global characteristics into the speech generating process.
\end{abstract}


\section{Introduction}
Natural human speech is very expressive, and varies based on the speaker individualities (such as age and gender), emotions and speaking styles (see, e.g., \cite{Erickson2005,Eyben2012}).
Many studies have suggested properly incorporating such expressiveness makes speech synthesis (SS) systems more pleasant to listen to and interact with, and have investigated the problem of expressive speech synthesis (see, e.g., \cite{Eyben2012,Charfuelan2013,Henter2017}).

This paper addresses the problem of synthesizing expressive speech without relying on speech expression labels, which we refer to as {\em unsupervised expressive speech synthesis }(UESS).
Many studies have reported that such labels are helpful for modeling complex audio data \cite{Fan2015,Luong2017,Henter2017,Lee2017}.
Unsupervised methods, however, are more desirable because expressive speech is easy to obtain from video hosting websites (e.g., Youtube) or audiobooks but annotating such sources is costly.
Moreover, manually-annotated labels are not always reliable: for example, not  all emotions in a given category have the same strength \cite{Henter2017}.

Another important aspect of this paper is that it focuses on the {\em neural autoregressive models}, which have been shown to offer significant performance improvements to SS systems. For example, WaveNet \cite{wavenet} generates more natural speech than traditional parametric or unit-selection based SS methods.
In addition, autoregressive-based sequence-to-sequence (seq2seq) speech synthesis models have simple structure, and can be trained on $<$text, audio$>$ pairs with minimal human annotation.
Such end-to-end systems have many advantages, for example, they alleviate the need for laborious feature engineering, which may involve heuristics, and are likely to be more robust than multi-stage models where each component's errors can compound \cite{tacotron}.

However, finding ways to add the expressiveness to autoregressive SS models is still an open issue.
One of the difficulties here is such models are typically unable to model the global characteristics of data because they model data densities autoregressively, i.e., point-by-point \cite{kolesnikov2017, pixelvae}.
Given that certain sources of speech expressiveness (e.g., gender or emotions) characterize speech in a global manner (sentence level), autoregressive SS models may suffer from the difficulty: it reduces the quality of the synthesized speech, and they have no structured way to control the expressions in the synthesized speech.

In this paper, we propose a model called VAE-Loop, which combines VoiceLoop \cite{voiceloop}, an autoregressive SS model, with Variational Autoencoer (VAE) \cite{VAE}.
Several recent studies (e.g., \cite{Hsu2017}) have shown VAEs can model global characteristics of speech such as speaker individualities, but to our knowledge no study has yet suggested using VAEs for SS or UESS.
We use VAE to deal with the problem incorporatig global characteristics into the speech generating process when using an autoregressive model for UESS.
Specifically, VAE encodes such global characteristics as a tractable probability distribution, which is used to give hints about them to VoiceLoop, allowing it to generate higher quality speech and to control the expressions in the synthesized speech.
The proposed VAE-based method is both effective and simple, trained in end-to-end as well as seq2seq SS models.

Our experiments show, by incorpolating global characteristics in this way, the VAE can help VoiceLoop to attain lower test errors and higher mean opinion scores (MOSs) when no labels are available.
Also, latent variables yielded by the VAE show it has the ability to control speaker individualities and speaking styles, and interpolate between them.

\section{Related Work}

Seq2seq SS systems have simple structures that directly predict acoustic features from text.
In addition, they have demonstrated the ability to generate natural and intelligible speech \cite{tacotron, char2wav, deepvoice3} and to robustly handle different prosodies \cite{voiceloop, Ronanki2017}.
\cite{voiceloop} has shown even when VoiceLoop is trained on data obtained from YouTube containing various speaking styles, it can generate high quality speech.
We have incorporated VoiceLoop into the proposed model, expecting that it will be effective for UESS.

Several studies tackled the problem of UESS on both neural autoregressive and non-neural paradigm.
On the non-neural paradigm, representation of emotions, acquired by unsupervised learning methods such as clustering and principal component analysis, have been used to generate speech \cite{Eyben2012, Charfuelan2013, Chen2014}.
The most relevant works to ours might be \cite{Wang2017,voiceloop}, which proposed the seq2seq SS models that can learn and control speaking styles in an unsupervised manner.
However, the proposed VAE-based method is different in that it learns speech expressions as a tractable distribution, which can be useful for downstream tasks such as interpolation and semi-supervised learning.

\cite{Eyben2012,Chen2014} pointed out that UESS can be divided into two parts: predicting expressive information from text; and synthesizing the speech with a particular expression.
In this paper only the latter stage is considered for simplicity.

Several recent studies have proposed using VAEs for modeling speech \cite{Blaauw2016, Hsu2016, Hsu2017, VQ-VAE, Hsu2017VC}.
The most relevant works might be \cite{Hsu2016,Hsu2017VC}, which conditioned the VAE on speaker labels and perform voice conversion.
In contrast, we perform SS by conditioning the model on text, and verify that the VAE as a SS model is also able to learn and control speech expressions.

Many other studies in areas outside the SS field have also proposed combining VAEs and autoregressive models.
For example, it has been shown a recurrent neural network language model combined with VAE can generate sentences with consistent global characteristics (e.g., style, topics) \cite{Samuel2016} .
That study pointed out also the issue that autoregressive models often ignore the latent variables obtained from the VAE.
Several authors have proposed countermeasures for dealing with this problem \cite{pixelvae, vlae, VQ-VAE}.
Based on these studies, we employ the {\em KL cost annealing} approach used in \cite{Samuel2016} to alleviate this problem.

\section{Models}

In this section, we first introduce conditional VAE and VoiceLoop, which form the basis of the proposed method, and then present our VAE-Loop method.

\subsection{Conditional Variational Autoencoder}

Here we present the variant of VAE used in VAE-Loop, which is simply conditioned on the auxiliary features $\bm{c}$.
In this study, $\bm{x}$ and $\bm{c}$ correspond to the acoustic features and phonemes, respectively.

Using the latent variable vector $\bm{z}$ and the approximate distribution $q_{\bm{\phi}}(\bm{z}|\bm{x}, \bm{c})$ (with parameter $\bm{\phi}$) of the true posterior $p_{\bm{\theta}}(\bm{z}|\bm{x}, \bm{c})$ (with parameter $\bm{\theta}$), we can obtain the following lower bound $\mathcal{L}(\bm{\theta}, \bm{\phi}; \bm{x}, \bm{c})$ on the marginal likelihood of $p_{\bm{\theta}}(\bm{x}|\bm{c})$:
\begin{align}
  \mathcal{L}(\bm{\theta}, \bm{\phi}; \bm{x}, \bm{c}) =& \int q_{\bm{\phi}}(\bm{z}|\bm{x}, \bm{c}) \log \frac{p_{\bm{\theta}}(\bm{x}, \bm{z}|\bm{c})} {q_{\bm{\phi}}(\bm{z}|\bm{x}, \bm{c})} dz \\
  =& \int q_{\bm{\phi}}(\bm{z}|\bm{x}) \log \frac{p_{\bm{\theta}}(\bm{x}, |\bm{z}, \bm{c}) p(\bm{z})} {q_{\bm{\phi}}(\bm{z}|\bm{x})} dz  \\
  = - D_{KL} (q_{\bm{\phi}}(\bm{z}|\bm{x})&||p(\bm{z})) + \EX_{q_{\bm{\phi}}(\bm{z}|\bm{x})}[\log p_{\bm{\theta}}(\bm{x}|\bm{z}, \bm{c}) ] \label{eq:cvae:ll}
\end{align}
where we have assumed $q_{\bm{\phi}}(\bm{z}|\bm{x}, \bm{c})=q_{\bm{\phi}}(\bm{z}|\bm{x})$ and $p(\bm{z}|\bm{c})=p(\bm{z})$ for simplicity.
The prior $p(\bm{z})$ and approximate posterior $q_{\bm{\phi}}(\bm{z}|\bm{x})$ are modeled by Gaussian distributions, namely $p(\bm{z}) = \mathcal{N}(\bm{z} | \bm{0}, \bm{I})$ and $q_{\bm{\phi}}(\bm{z}|\bm{x}) = \mathcal{N}(\bm{z} | \bm{\mu}_{\bm{\phi}}(\bm{x}), \bm{\sigma}^2_{\bm{\phi}}(\bm{x}) \bm{I})$.

During training, we update the parameters $\bm{\theta}$ and $\bm{\phi}$ to maximize $\mathcal{L}$.
We call $q_{\bm{\phi}}(\bm{z}|\bm{x})$ an encoder and $p_{\bm{\theta}}(\bm{x}|\bm{z}, \bm{c})$ a decoder.

\subsection{VoiceLoop}

Let $\bm{x} = [\bm{x}_1, ..., \bm{x}_T]$ be a variable length sequence of audio features that we want to predict.
VoiceLoop can be regarded as a conditional autoregressive model with a parameter $\bm{\xi}$ as follows:
\begin{align}
  p_{\bm{\xi}}(\bm{x}|\bm{c})&=\prod_{t=1}^{T} p_{\bm{\xi}}(\bm{x}_t|\bm{x}_{1:t-1}, \bm{c}) \label{eq:voiceloop:l} \\
  p_{\bm{\xi}}(\bm{x}_t|\bm{x}_{1:t-1}, \bm{c}) &= \mathcal{N}(\bm{x}_t | \bm{\mu}_{\bm{\xi}}(\bm{x}_{1:t-1}, \bm{c}), \bm{I}) \label{eq:voiceloop:gauss}
\end{align}
where $\bm{x}_{1:t-1}$ is the audio features between time steps $1 \; and \; t-1$ and we estimate $p_{\bm{\xi}}(\bm{x}_t|\bm{x}_{1:t-1}, \bm{c})$ in order for each time step $t \in T$ .
Eq.(\ref{eq:voiceloop:gauss}) assumes $p_{\bm{\xi}}(\bm{x}_t|\bm{x}_{1:t-1}, \bm{c})$ is modeled by a Gaussian distribution with mean $\bm{\mu_\xi}$ and variance $\bm{I}$ (identity matrix).

Next, we describe the procedure for estimating $p_{\bm{\xi}}(\bm{x}_t|\bm{x}_{1:t-1}, \bm{c})$.
VoiceLoop has a shifting buffer, which can be seen as a matrix $S \in \mathbb{R}^{d \times k}$ with columns $S[1]...S[k]$. At each time step, all the columns shift to the right as follows:
\begin{align}
  S_t [i+1] &= S_{t-1} [i] \;\; for \;\; 1 \leq i < k \label{eq:voiceloop:buffer1} \\
  S_t [1] &= u \label{eq:voiceloop:buffer2}
\end{align}
Here, $u$ is a function of four parameters, namely the current attention-mediated context $\bm{c}_t$, buffer $S_{t-1}$ itself, latest ``spoken'' output $\bm{x}_{t-1}$ and speaker embedding $s$, as follows:
\begin{align}
  C_t =& [\bm{c}_t + tanh(F_u (s)), \bm{x}_{t-1} ] \label{eq:voiceloop:C} \\
  u =& N_u ([S_{t-1}, C_t]) \label{eq:voiceloop:u}
\end{align}
where $[a, b]$ is the concatenation of the two column vectors $a$ and $b$ to one column vector.
VoiceLoop then estimates $\bm{x}_{t}$ using the buffer $S_t$ and embedding $s$, as follows:
\begin{align}
  \hat{\bm{x}}_t = N_o (S_t + F_o (s)) \label{eq:voiceloop:xhat}
\end{align}
where $F_u, N_u, F_o$ and $N_o$ are the respective neural networks, and $\hat{\bm{x}}_t$ is equivalent to $\bm{\mu}_{\bm{\xi}}$ in Eq.(\ref{eq:voiceloop:gauss}).

\subsection{Proposed model: VAE-Loop}

VoiceLoop has no structured way to model the complex global characteristics in an unsupervised manner since it relies on the point-by-point estimation.
In contrast, VAE-Loop explicitly incorporates them into the speech generating process in the VAE framework.
Specifically, VAE-Loop regards VoiceLoop as a decoder for the conditional VAE,
i.e., VoiceLoop is conditioned on the global latent variable $\bm{z}$.

\subsubsection{Modeling various expressions using VAE}

We first change VoiceLoop's probability distribution (Eq.(\ref{eq:voiceloop:l}) and (\ref{eq:voiceloop:gauss})) so that it is conditioned on the latent variable $\bm{z}$, as follows:
\begin{align}
  p_{\bm{\theta}}(\bm{x}|\bm{z}, \bm{c})&=\prod_{t=1}^{T} p_{\bm{\theta}}(\bm{x}_t|\bm{x}_{1:t-1},\bm{z}, \bm{c}) \label{eq:proposed:l} \\
  p_{\bm{\theta}}(\bm{x}_t|\bm{x}_{1:t-1},\bm{z}, \bm{c}) &= \mathcal{N}(\bm{x}_t | \bm{\mu}_{\bm{\theta}}(\bm{x}_{1:t-1}, \bm{z}, \bm{c}), \bm{I}) \label{eq:proposed:gauss}
\end{align}
where we have set $\bm{\xi}=\bm{\theta}$ because VoiceLoop is regarded as the decoder in Eq.(\ref{eq:cvae:ll}).

In the VAE framework, information which is useful to estimate $\bm{x}$ but is not contained in the text $\bm{c}$ is encoded into $\bm{z}$.
Since certain types of expressions are difficult to predict from the spoken text alone, $\bm{z}$ is expected to acquire latent representations of such expressions (i.e., the global characteristics).

\subsubsection{Generating speech using the global latent variable}
The $p_{\bm{\theta}}(\bm{x}_t|\bm{x}_{1:t-1},\bm{z}, \bm{c})$ in Eq.(\ref{eq:proposed:gauss}) is estimated by changing Eq.(\ref{eq:voiceloop:u}) to incorporate the latent variable $\bm{z}$ into the speech generating process of VoiceLoop, as follows:
\begin{align}
  u = N_u ([S_{t-1}, C_t, \bm{z}]) \label{eq:proposed:u}
\end{align}
As previously mentioned, $\bm{z}$ is expected to acquire the expression information.
In addition, since $\bm{z}$ does not depend on the time step $t$ unlike $S_{t-1}$ and $C_t$, it conditions the speech generating process in a global manner.

\subsubsection{Training and inference}

By combining Eq.(\ref{eq:cvae:ll}) and (\ref{eq:proposed:l}), we can obtain the objective function of VAE-Loop:
\begin{align}
  \mathcal{L}(\bm{\theta}, \bm{\phi}; \bm{x}, \bm{c}) =& - D_{KL} (q_{\bm{\phi}}(\bm{z}|\bm{x})||p(\bm{z}))  + \nonumber \\
  &\EX_{q_{\bm{\phi}}(\bm{z}|\bm{x})}[\sum_{t=1}^{T} \log p_{\bm{\theta}}(\bm{x}_t|\bm{x}_{1:t-1},\bm{z}, \bm{c}) ] \label{eq:proposed:ll2}
\end{align}
where the first and second terms are the regularizer and reconstruction error, respectively.
We can estimate the reconstruction error by taking the mean squared error between the estimators $\hat{\bm{x}}_t$ and true audio features $\bm{x_t}$.
The second term is thus equivalent to the objective function of the original VoiceLoop, except that $\bm{z}$ is used in the generating process.

\begin{figure}[t]
  \centering
  \includegraphics[width=8cm]{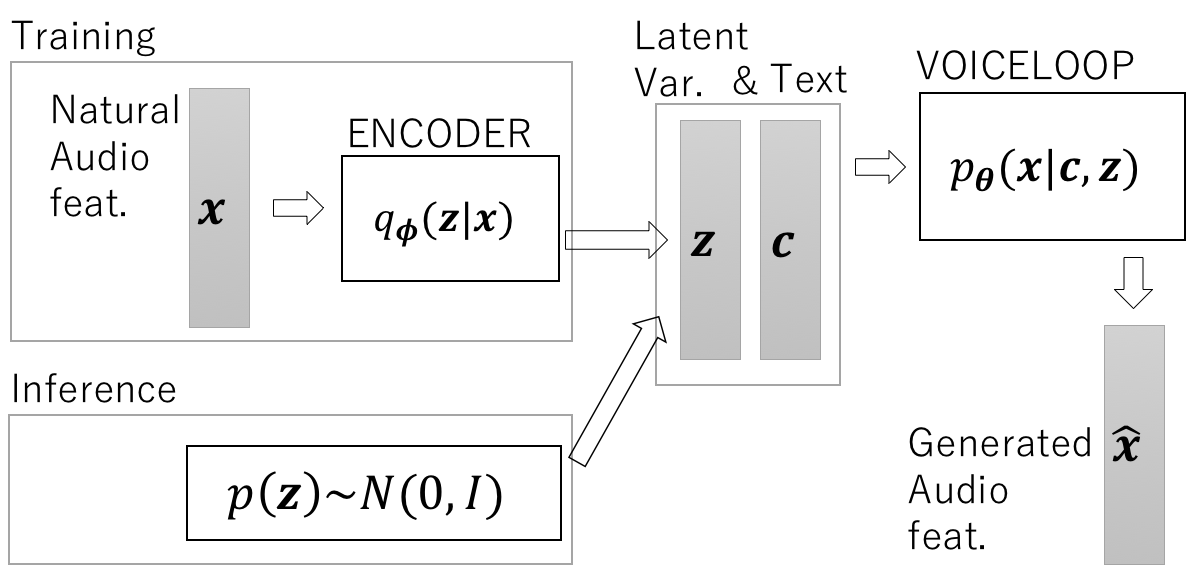}
  \caption{Speech generating process of VAE-Loop}
  \label{image:model}
\end{figure}

At training, $\bm{z}$ is sampled from the encoder.
Here, as with a conventional VAE, the encoder $q_{\bm{\phi}}(\bm{z}|\bm{x})$ is parameterized as a deep neural network (DNN).
At inference, $\bm{z}$ is sampled from the prior $p(\bm{z})$.
Figure \ref{image:model} illustrates the speech generating process of VAE-Loop,
showing its training and inference procedures are simple; and do not not require any additional training stage or data preprocessing compared with VoiceLoop alone.
In addition, in spite of these simple procedures, it offers higher performance as we will demonstrate in Section 4.

\subsubsection{KL cost annealing}

We exploit the ideas outside the SS field and employ the simple {\em KL cost annealing} technique to alleviate the problem that autoregressive models often ignore the latent variables \cite{Samuel2016}.
They argued that the latent variables were ignored because the regularizer in Eq.(\ref{eq:proposed:ll2}), which we call Kullback-Leibler divergence (KLD) term, acted too strongly at the start of training; therefore, Eq.(\ref{eq:proposed:ll2}) is adjusted to include the weight $\lambda$, as follows:
\begin{align}
  \mathcal{L}(\bm{\theta}, \bm{\phi}; \bm{x}, \bm{c}) =& - \lambda D_{KL} (q_{\bm{\phi}}(\bm{z}|\bm{x})||p(\bm{z}))  + \nonumber \\
  & \EX_{q_{\bm{\phi}}(\bm{z}|\bm{x})}[\sum_{t=1}^{T} \log p_{\bm{\theta}}(\bm{x}_t|\bm{x}_{1:t-1},\bm{z}, \bm{c}) ] \label{eq:proposed:ll3}
\end{align}
We set $\lambda$ to 0 at the start of training so that the model learns to encode as much information as possible, and then increase it linealy to 1 over the course of the annealing process.

\section{Experiments}

\subsection{Datasets}

We used two datasets: one featuring multiple speakers and another containig a variety of emotions and speaking styles.
The first was VCTK Corpus\cite{VCTK} (VCTK), which contains speech samples from 109 English speakers.
We used the version of the dataset from VoiceLoop's source code page\footnote{\url{https://github.com/facebookresearch/loop/}} instead of the complete VCTK, in order to replicate the conditions of \cite{voiceloop}.
This contained about 5 hours of speech by 21 North American speakers (4 males and 17 females), and each utterance lasted less than 5 seconds.
Our second dataset was from the Blizzard Challenge 2012 (Blizzard2012), and consisting of four audiobooks \cite{Blizzard2012,Braunschweiler2010}.
Audiobooks are often used by expressive speech synthesis studies because they include a variety of emotions and speaking styles.
Unlike those in the VCTK, all the utterances in this dataset were read by the same male speaker.
To match the conditions with VCTK and avoid exploding gradients, we used utterances of less than 5 seconds only, resulting in a total of about 10 hours of speech.
Both of the datasets were divided into three parts, with 90\% used for training and the remaining 10\% used for validation, with 50 samples set aside as test data.

\subsection{Experimental setup}

We used DNN based on time-domain convolution for the encoder of VAE-Loop.
Specifically, the first half of the encoder consisted of five repeated convolutional layers with a stride size 2, with dropout, batch normalization and ReLU, while the rest consisted of time-domain global max-pooling and fully connected layers.
The model hyperparameters used for the baseline VoiceLoop and VAE-Loop decoder were the same as in the authers' implementation$^1$.

During training, we employ a variant of teacher forcing technique as well as the original VoiceLoop \cite{voiceloop}, which aims to stabilize both training and inference.
We refer to this as {\em semi-teacher-forcing}.
Specifically, $\bm{x}_{t-1}$ used as input to the network $N_u$ in Eq.(\ref{eq:voiceloop:C}) is replaced by $\tilde{\bm{x}}_{t-1}$ bellow:
\begin{align}
  \tilde{\bm{x}}_{t-1} = \frac{\bm{x}_{t-1} + \hat{\bm{x}}_{t-1}} {2} + \eta \label{eq:semi-teacherforce}
\end{align}
where we assumed $\eta \sim \mathcal{N}(0, \bm{I})$.

Unless otherwise noted, we used the Adam \cite{adam} optimizer and 150 training epochs.
The learning rate was chosen to achieve the minimum possible validation error from the set [1e-3, 1e-4, 5e-5, 1e-5], resulting in 1e-4 and 5e-5 for the VCTK and Blizzard2012, respectively.

\subsection{Effect of latent variables on the test error}

{\setlength\floatsep{0pt}
\setlength\abovecaptionskip{5pt}
\setlength\textfloatsep{0pt}
\setlength\intextsep{5pt}
\begin{table}[t]
  \caption{Test errors for different numbers of annealing epochs and $\bm{z}$ dimensions, on the VCTK dataset.}
  \label{table:loss}
  \centering
  \scalebox{0.85}{
    \begin{tabular}{llllll}
      \hline
      Model & annealing & z-dim & Rec. & KLD  & Total \\
            & epochs &          & error & term \\
      \hline \hline
      VoiceLoop (w/o) & N/A & N/A   & 15.946 & N/A & \bf{15.946} \\
      VoiceLoop (w/)  & N/A & N/A   & 15.759 & N/A & 15.759 \\
      \hline
      VAE-Loop             & 0        & 64  & 15.832 & 0.073 & 15.905 \\
      VAE-Loop             & 15(10\%) & 64  & \bf{15.684} & \bf{0.090} & \bf{15.774} \\
      VAE-Loop             & 30(20\%) & 64  & 15.749 & 0.086 & 15.835 \\
      VAE-Loop             & 15       & 32  & 15.839 & 0.082 & 15.921 \\
      VAE-Loop             & 15       & 128 & 15.724 & 0.084 & 15.808 \\
      \hline
    \end{tabular}
  }
\end{table}
}

We compared the test errors for VAE-Loop with those for VoiceLoop alone, to demonstrate how adding latent variables to VoiceLoop enables it to estimate audio features more accurately.
For this experiment, the models were trained on the VCTK, using the setup described in Section 4.2.
However, to stabilize training on various hyperparameters, we set the learning rate to 5e-5.
In addition, since the baseline VoiceLoop had not converged sufficiently after 150 epochs at that learning rate, we extended the training period to 200 epochs.
We tested with annealing for 10 or 20\% of the training period, and without annealing.
The test errors were calculated using semi-teacher-forcing in order to use the same objective function as during training.

Table \ref{table:loss} presents the test errors, calculated using Eq.(\ref{eq:proposed:ll2}) and then divided by the sequence length. Here, (w) and (w/o) mean ``with speaker labels'' and ``without speaker labels'' respectively.
These show proper use of KL cost annealing leads to a higher KLD term and a lower test error, suggesting it allows the decoder (VoiceLoop) to recieve more useful information from the latent variables.
Moreover, the test errors of VAE-loop is smaller than that of VoiceLoop without speaker labels,
suggesting incorporating latent variables into the speech generating process enables VoiceLoop to estimate audio features more accurately.

\subsection{Mean opinion score tests}

{\setlength\textfloatsep{0pt}
\setlength\abovecaptionskip{5pt}
\setlength\textfloatsep{0pt}
\setlength\intextsep{5pt}
  \begin{table}[t]
    \caption{Mean opinion scores (mean ± CI) for both datasets.}
    \label{table:mos}
    \centering
    \scalebox{1}{
      \begin{tabular}{lrr}
        \hline
        Method  & VCTK &  Blizzard2012  \\
        \hline \hline
        Ground Truth & 4.07 $\pm$ 0.23 & 3.94 $\pm$ 0.30 \\
        \hline
        VoiceLoop(w/o)  & 2.51 $\pm$ 0.34 & 2.23 $\pm$ 0.24\\
        VoiceLoop(w/)   & 3.24 $\pm$ 0.27 & N/A \\
        VoiceLoop(orig, w/) & 3.57            & N/A \\
        \hline
        VAE-Loop($\sigma=1$)   & \bf{3.25 $\pm$ 0.29} & 2.47 $\pm$ 0.32 \\
        VAE-Loop($\sigma=0.7$) & N/A & \bf{2.89 $\pm$ 0.32} \\
        VAE-Loop($\sigma=0$)   & N/A & \bf{3.03 $\pm$ 0.32} \\
        \hline
      \end{tabular}
    }
  \end{table}
}

To demonstrate that incorporating global characteristics enables VAE-Loop to generate higher quality speech, we conducted an mean opinion score (MOS) study, using the crowdMOS toolkit \cite{crowdmos} and Amazon Mechanical Turk.
The MOS is a popular subjective audio quality measure, obtained by asking people to rate the audio's naturalness on a scale of 1 to 5.
More than 15 people living in the US rated each of the two datasets.
Table \ref{table:mos} shows MOSs for the two models, together with their 95\% confidence intervals(CIs).
Here, ``Ground Truth'' recordings were the audio reconstructed using the WORLD vocoder \cite{WORLD}.

For the VCTK, the MOS achieved by VAE-Loop was higher than that by VoiceLoop without speaker labels, matching even VoiceLoop with labels, despite not using labels.
In addition, in our informal listening tests, we observed VAE-Loop was less likely than the baseline to generate unintelligible speech (e.g., several seconds of just breath or a certain phoneme).
Therefore, these results could indicate that where the original VoiceLoop struggled to model the various speaker individualities, adding VAE stabilized its speech generating process by giving hints about them.
Here, we acknowledge that VoiceLoop's MOS by our inplementation is lower than that reported in \cite{voiceloop} (``VoiceLoop(orig, w/)'' in Table \ref{table:mos}), probably because there might be a different choice of hyperparameters, including the use of pre-training.

For the Blizzard2012, we observed that the high variance of the $p(\bm{z})$ used for generating test samples meant VAE-Loop often generated unintelligible speech in much the same way as VoiceLoop. To investigate this issue, we instead assumed that $p(\bm{z}) = \mathcal{N}(\bm{z} | \bm{0}, \sigma^2 \bm{I})$ at inference time, and sampled $\bm{z}$ using different parameters $\sigma$, where $\sigma=0$ means we always sample $\bm{z}=\bm{0}$.
When the variance of $p(\bm{z})$ was suppressed, in this way, VAE-Loop's MOS improved, exceeding that of the baseline.
Here, note that using small $\sigma$ values means always sampling similar $\bm{z}$ values; therefore, VAE-Loop can make a tradeoff between stable inference and latent variable variety.

\subsection{Controlling speech expressions using latent variables}
To demonstrate that VAE-Loop is able to control the experssions in its synthesized speech, we presented the trajectories of the fundamental frequency (F0).
Figure \ref{image:vctk_f0} shows F0 trajectories generated by VAE-Loop, trained on the VCTK.
The left and right figures correspond to different texts; however, both were generated using the same $\bm{z}$ values.
Here, different latent variable values, $\bm{z_1}$ and $\bm{z_2}$, lead to different F0 characteristics, indicating our model can control speaker individualities expressed in the sythesized speech using latent variables.
Moreover, when the speech is synthesized using a latent variable value that interpolated between previous two, the F0 trajectories were also averaged.
Likewise, Figure \ref{image:blizzard_f0} shows F0 trajectories generated by VAE-Loop, trained on the Blizzard2012. Here, the latent variables characterize the pitch fluctuations of the F0 trajectories.
Some audio samples can be found at: \\ \url{https://akuzeee.github.io/VAELoopDemo/}.

{\setlength\floatsep{0pt}
\setlength\abovecaptionskip{0pt}
\setlength\textfloatsep{0pt}
\setlength\intextsep{5pt}
\begin{figure}[t]
  \centering
  \includegraphics[width=8cm]{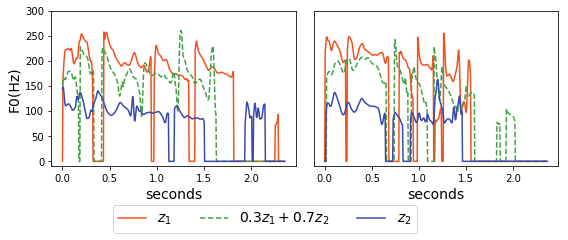}
  \caption{F0 Trajectories for two utterances generated by VAE-Loop, trained on VCTK. Here, $\bm{z}_1$ and $\bm{z}_2$ correspond to high-pitched (female as we heard) and low-pitched (male) voices, respectively. Averaged F0 trajectories are also shown, generated by interpolating between $\bm{z}_1$ and $\bm{z}_2$.}
  \label{image:vctk_f0}
\end{figure}
\begin{figure}[t]
  \centering
  \includegraphics[width=8cm]{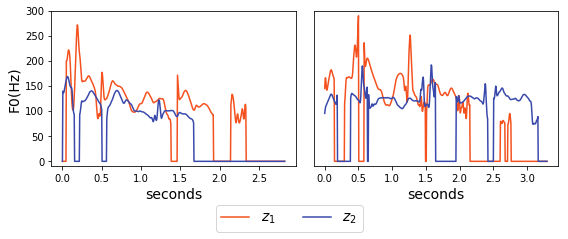}
  \caption{F0 Trajectories for two utterances generated by VAE-Loop, trained on Blizzard2012. Here, $\bm{z}_1$ and $\bm{z}_2$ correspond to voices with large (dramatic as we heard) and small (calm) pitch fluctuations, respectively.}
  \label{image:blizzard_f0}
\end{figure}
}

\section{Conclusions}

In this paper, we have proposed to combine VoiceLoop with VAE, in order to enable this autoregressive SS model to be more expressive by using VAE to help model a range of expressions.
Even though autoregressive SS models have shown promising results, they typically lack the ability to model the global characteristics of speech.
However, the proposed method can incorporate such expressions explicitly into the speech generating process in an unsupervised manner.
Our experiments have shown taking advantage of these global characteristics could enable our method to generate higher quality speech than VoiceLoop without labels and to control speech expressions.

In future studies, we plan to extend this approach to semi-supervised learning with a small amount of labeled data, and to infer the latent variables from text.

% Generated by IEEEtran.bst, version: 1.13 (2008/09/30)

\end{document}